\def\year{2022}\relax
\documentclass[letterpaper]{article} 
\usepackage{aaai22}  

\usepackage{times}  
\usepackage{helvet}  
\usepackage{courier}  
\usepackage[hyphens]{url}  
\usepackage{graphicx} 
\def\UrlFont{\rm}  
\usepackage{natbib}  
\usepackage{caption} 
\DeclareCaptionStyle{ruled}{labelfont=normalfont,labelsep=colon,strut=off} 
\usepackage{booktabs}
\newenvironment{smallarray}[1]
 {\null\,\vcenter\bgroup\scriptsize
  \arraycolsep=.13885em
  \hbox\bgroup$\array{@{}#1@{}}}
 {\endarray$\egroup\egroup\,\null}
\usepackage[ruled,linesnumbered]{algorithm2e}

\usepackage{newfloat}
\usepackage{listings}
\nocopyright
\usepackage{multirow}

\title{PTQ-SL: Exploring the Sub-layerwise Post-training Quantization}
\author{
    Zhihang Yuan, \textsuperscript{\rm 1,2}
    Yiqi Chen, \textsuperscript{\rm 1}
    Chenhao Xue, \textsuperscript{\rm 1} \\
    Chenguang Zhang, \textsuperscript{\rm 1,2}
    Qiankun Wang, \textsuperscript{\rm 1,2}
    Guangyu Sun, \textsuperscript{\rm 1}.
}
\affiliations{
    \textsuperscript{\rm 1}Center for Energy-Efficient Computing and Applications, Peking University, Beijing, China \\
    \textsuperscript{\rm 2} Houmo AI, Beijing, China \\
    \{yuanzhihang, gsun\}@pku.edu.cn

}

\usepackage{bibentry}

\begin{document}

\maketitle

\begin{abstract}

Network quantization is a powerful technique to compress convolutional neural networks. 
The quantization granularity determines how to share the scaling factors in weights, which affects the performance of network quantization.
Most existing approaches share the scaling factors layerwisely or channelwisely for quantization of convolutional layers. Channelwise quantization and layerwise quantization have been widely used in various applications.
However, other quantization granularities are rarely explored.
In this paper, we will explore the sub-layerwise granularity that shares the scaling factor across multiple input and output channels.
We propose an efficient post-training quantization method in sub-layerwise granularity (PTQ-SL).
Then we systematically experiment on various granularities and observe that the prediction accuracy of the quantized neural network has a strong correlation with the granularity.
Moreover, we find that adjusting the position of the channels can improve the performance of sub-layerwise quantization.
Therefore, we propose a method to reorder the channels for sub-layerwise quantization.
The experiments demonstrate that the sub-layerwise quantization with appropriate channel reordering can outperform the channelwise quantization. 

\end{abstract}

\section{Introduction}

In the past decade, the neural network has achieved great success and changed our way of life and industrial production.
The success comes mainly from the increasing scale of the neural network, which consumes more and more memory and computation resources.
The large scales of modern neural networks hinder the applications, such as healthcare monitoring and autonomous driving, which require real-time inference on resource-constrained hardware.
It is essential to compress the neural network to meet the constraints of inference latency, memory footprint, and energy consumption.

Network quantization is one of the most widely used techniques to compress the neural network.
It transforms the data in the neural network from floating-point values to integer values with lower bit-width.
In this way, avoiding expensive floating-point computation increases the inference speed. 
And the memory footprint and energy consumption are also reduced.
However, the cost of network quantization is the prediction accuracy drop.
Therefore, various methods have been proposed to increase the accuracy of the quantized network, such as fine-tuning the network~\cite{Quantization_training_integer_only_cvpr2018,QIL_cvpr2019}, using dynamic quantization of activation~\cite{PTQ_4bit_rapid_deployment_nips2019,qpp_real_time_quantization_cvpr2021}, and bias correction~\cite{DFQ_bias_correction_iccv2019,Adaquant_arxiv2020}. 

\begin{figure}[tb] 
    \centering
    \includegraphics[width=0.39\textwidth]{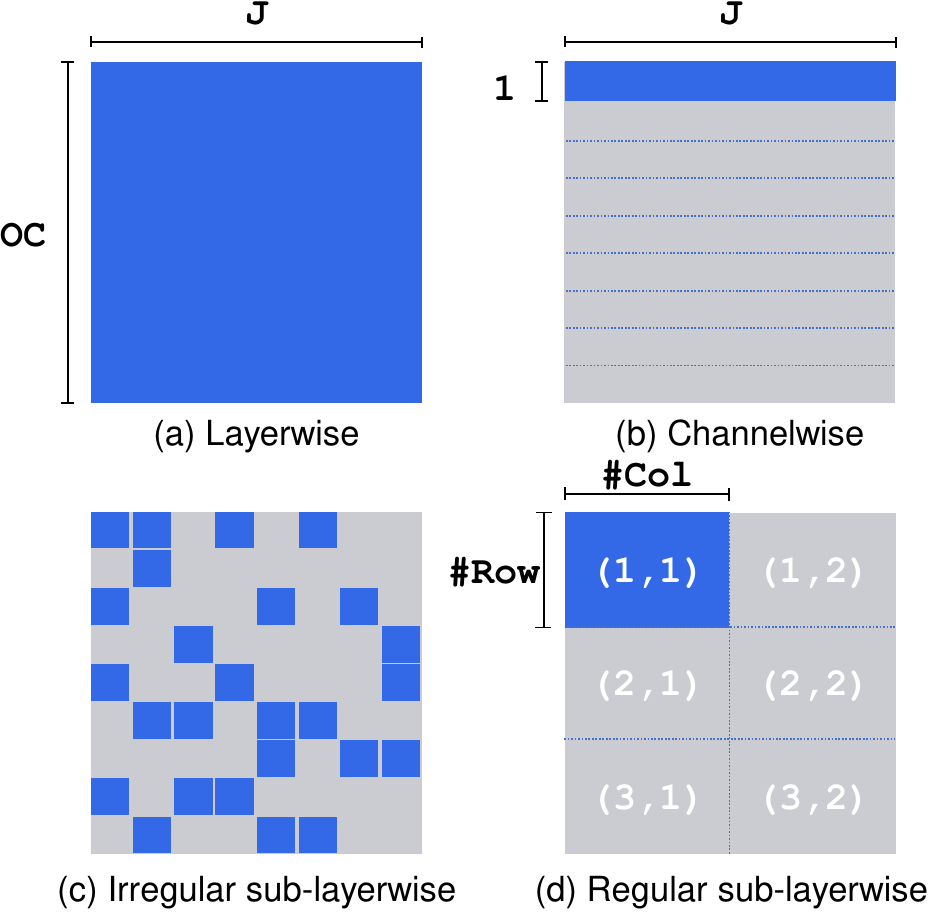} 
    \caption{Different granularities for weight quantization. The weights in the blue area share the same quantization parameter.}
    \label{im_granularities}
\end{figure}

The quantization granularity determines how the quantization parameters are shared among the weights, which has a great influence on the quantization performance.
But it has not been fully explored.
In respect of the granularity, existing methods are categorized into two types,  which are layerwise quantization and
sub-layerwise quantization.
As shown in Figure~\ref{im_granularities}, layerwise quantization uses a single parameter to quantize the weights of a layer, while sub-layerwise quantization shares a parameter in each weights subset.
Because the distributions of the weights in different subsets are different, the fine-grained quantization treats them with different quantization parameters.
Thus, sublayerwise quantization can achieve higher performance than the layerwise approach. 
The channelwise quantization is a special case of the sub-layerwise quantization for convolutional neural networks (CNNs), which has been widely proved to be effective in practice~\cite{Survey_quantization_arxiv2021}.


In our opinion, the channel granularity might not be the best choice.
Other sub-layerwise granularities are the potential to improve the quantization performance.
Recently, we notice that the processing-in-memory-based hardware accelerators have developed rapidly for network inference~\cite{Survey_RRAM_PIM_2019, Compression_hardware_acc_survey_2020}.
They split the matrix multiplication into multiple sub-matrix multiplications and process them separately as shown in Figure~\ref{split_matrix_multiplication}.
So the sub-layerwise quantization is required by these accelerators.
However, the sub-layerwise quantization is rarely explored.
In this paper, we will systematically study the sub-layerwise granularity for weight quantization.

We propose an efficient post-training quantization method in sub-layerwise granularity (PTQ-SL) and explore various configurations of granularity.
We observe that the prediction accuracy of the quantized neural network has a strong correlation with the granularity.
Moreover, we find that the performance of sub-layerwise quantization can be improved by adjusting the position of the channels.
Therefore, we propose an evolutionary algorithm-based method to reorder the channels for sub-layerwise quantization.
By joint reordering the channels in adjacent layers, the channel reordering will not decrease the inference speed.
We evaluate the method on various computer vision tasks including image classification and object detection.
At last, we discuss the computation overhead and the memory overhead of the sub-layerwise quantization.
Considering the performance and insignificant overhead, sub-layerwise quantization can be a better choice than channelwise quantization.

\section{Sub-layerwise Quantization}
In this section, we first formulate the symmetric uniform quantization.
Then, we introduce different quantization granularities.
At last, we propose the sub-layerwise quantization method.

\subsection{Symmetric Uniform Quantization}
Conventionally, the weights and the activations in neural networks are single-precision floating-point values.
The goal of $k$-bit quantization is to transform each 32-bit floating-point value to $k$-bit integer value ($k<32$).
A mapping function is used to cast the original floating-point value space to the $k$-bit integer value space.

As with most of the previous work, we use the symmetric uniform mapping function $Q_k$. 
\begin{equation}
    Q_k(x,\Delta_x)=\mathit{clamp}(\mathit{round}(\frac{x}{\Delta_x}),-2^{k-1},2^{k-1}-1)
\end{equation}
, where $x$ is floating point value, the scaling factor $\Delta_x$ is the quantization parameter, and $Q_k(x,\Delta_x)$ is the converted $k$-bit integer value. 
The $\mathit{round}$ function casts the scaled floating-point value $\frac{x}{\Delta_x}$ to the nearest integer value, and the $\mathit{clamp}$ operation limits output to $k$-bit signed integer range $[-2^{k-1},2^{k-1}-1]$.
For example, the range of 8-bit mapping function $Q_8$ is $[-128,127]$. 
In this way, the original floating-point value can be approximated by re-scaling the quantized value:
\begin{equation}
    x\approx \Delta_x Q_k(x,\Delta_x)
\end{equation}

\subsection{Quantization Granularity}
To quantize a network, we should assign each weight with the quantization parameter, which is the scaling factor in symmetric uniform quantization.
Since there are a lot of weights in a neural network, it is necessary that a group of weights shares a single scaling factor to reduce the memory footprint and the computation overhead.
The quantization granularity defines which weights share a scaling factor.

Since convolution is the main component in CNNs, we focus on the quantization granularity for the convolutional layer.
We start with formulating the convolution operation, which can be transformed to matrix multiplication using the image to column (im2col).
The weights and the input feature maps in the $l$-th layer are transformed to 2-D matrices $W^{l}\in R^{\mathit{OC} \times J}$ and $X^{l}\in R^{J \times P}$, respectively.
$\mathit{OC}$ is the number of output channels, $J$ is the number of weights for one output channel, and $P$ is the number of pixels in an output feature map.
The convolution operation is formulated as follows.
\begin{equation}
\label{conv_operation}
    X^{l+1}_{cp}=f(\sum_j W_{cj}^{l}X_{jp}^{l})
\end{equation}
, where $X^{l+1} \in R^{\mathit{OC}\times P}$ is the output feature maps and $f$ is the activation function.

\textbf{Layerwise quantization} quantizes all weights in a layer with a single scaling factor $\Delta_W^l$.
All input feature maps share another scaling factor $\Delta_X^l$.
The operation in Eq~\ref{conv_operation} can be approximated by Eq~\ref{layerwise_quantization} using layerwise quantization~\footnote{We ommit $k$ and $\Delta$ in the mapping function for simplicity.}.
\begin{equation}
\label{layerwise_quantization}
    \sum_k W_{cj}^{l}X_{jp}^{l}\approx \Delta_{W}^l \Delta_{X}^l \sum_j Q(W_{cj}^{l})Q(X_{jp}^{l})
\end{equation}

\begin{figure}[tb] 
    \centering
    \includegraphics[width=0.47\textwidth]{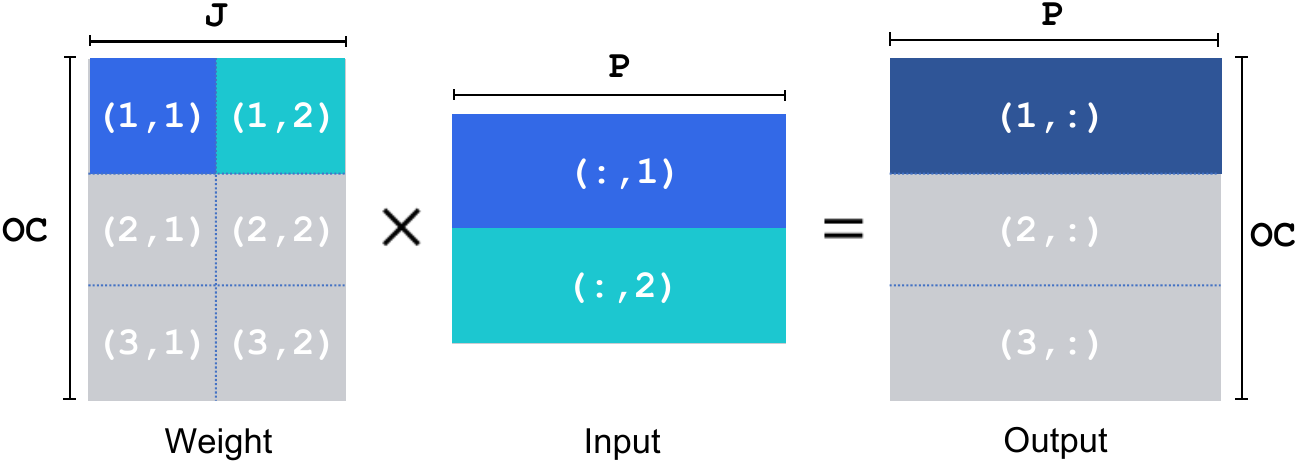} 
    \caption{The sub-matrix multiplication.}
    \label{split_matrix_multiplication}
\end{figure}

\textbf{Sub-layerwise quantization} uses multiple scaling factors to quantize a layer and shares each scaling factor among a subset of the weights.
As shown in Figure~\ref{im_granularities}, there are different types of sub-layerwise quantization granularities.
The irregular sub-layerwise quantization shares a scaling factor among arbitrary weights.
However, we should store the indexes of these weights, which brings the memory overhead. 
And we should use sparse computation, increasing the computation overhead.
To overcome these problems, the regular sub-layerwise quantization shares a scaling factor in a group of continuous weights.
This granularity contains $\#\mathit{Row}\times \#\mathit{Col}$ weights, where $\#\mathit{Row}$ is the number of rows and $\#\mathit{Col}$ is the number of columns.
In this way, the weight matrix can be divided into $\#V\times\#H$ sub-matrices.
We denote the $v$-th vertically and $h$-th horizontally sub-matrix $W^{l(v,h)}$ and its scaling factor $\Delta_W^{l(v,h)}$.
The computation of the matrix multiplication is formulated as follows~\footnote{Because we focus on the quantization of weight, we share a scaling factor among all inputs.}.
\begin{equation}
\label{sub_layerwise_computation}
    X^{l+1}_{cp}=f(\sum_{h}\Delta_{W}^{l(v,h)} \Delta_{X}^l \sum_{j} Q(W^{l(v,h)}_{cj})Q(X_{jp}^{l(:,h)}))
\end{equation}
where $X_{jp}^{l(:,h)}$ is the corresponding input sub-matrix, $h$ is in $[1,\#H]$ and $j$ is in $[1,\#\mathit{Col}]$.
Figure~\ref{split_matrix_multiplication} demonstrated this process.

\textbf{Channelwise quantization} is a special case of the regular sub-layerwise quantization with $\#\mathit{Col}=J$ and $\#\mathit{Row}=1$.
It quantizes the weights for different output channels of convolutional layers with different scaling factors. 
Since it can effectively improve the quantization performance, channelwise quantization has been widely used in various quantization methods.

\subsection{Sub-layerwise Post-training Quantization}
The layerwise quantization and the channelwise quantization have been well explored by previous methods.
However, the other sub-layerwise granularities are rarely explored.
To systematically explore the sub-layerwise quantization, we propose an efficient post-training quantization method for sub-layerwise granularity (PTQ-SL).

\begin{algorithm}[tb]
  \SetAlgoLined
  \caption{Iterative search algorithm for scaling factors of the weight sub-matrices}
  \label{iterative_search_scale}
  initialize $\Delta_{W}^{l(v,h)}\leftarrow \frac{\mathit{max}(\mathit{abs}({W}^{l(v,h)}))}{2^{k-1}}$\;
  initialize $\Delta_\mathit{opt}^{l(v,h)}\leftarrow \Delta_{W}^{l(v,h)}$\;
  \For{$iter$ = $1$ to $\#\mathit{Iter}$}{
    \For{$(v,h)$ = $(1,1)$ to $(\#V,\#H)$}
    {
      calculate $X^{l+1}$ with Eq~\ref{sub_layerwise_computation}\;
      $d\leftarrow \mathit{distance}(X^{l+1},X^{l+1}_\mathit{fp})$\;
      \For{$\Delta_\mathit{candidate}$ in $\mathit{space}(\alpha,\beta,\Delta_{W}^{l(v,h)})$}{
        $\Delta_{W}^{l(v,h)}\leftarrow \Delta_\mathit{candidate}$\;
        calculate $X^{l+1}$ with Eq~\ref{sub_layerwise_computation}\;
        \If{$\mathit{distance}(X^{l+1},X^{l+1}_\mathit{fp})<d$}{
          $\Delta_\mathit{opt}^{l(v,h)}\leftarrow \Delta_\mathit{candidate}$\;
          $d\leftarrow \mathit{distance}(X^{l+1},X^{l+1}_\mathit{fp})$\;
        }
      }
      $\Delta_{W}^{l(v,h)}\leftarrow \Delta_\mathit{opt}^{l(v,h)}$
    }
  }
\end{algorithm}

The goal of the quantization method is to determine the scaling factors of symmetric uniform mapping function for weights and feature maps.
Post-training quantization (PTQ) is widely used when the training dataset is unavailable or the computation resources are limited.
Like previous work~\cite{EasyQuant_arxiv2020}, we use some unlabeled input images to calibrate the scaling factors to minimize the quantization error layer-by-layer.
We define the quantization error as the Euclidean distance or cosine distance between the output feature maps of the quantized network $X^{l+1}$ and that of the original network $X^{l+1}_{fp}$.
The problem is formulated as follows.
\begin{equation}
\label{eq_distance}
    \min_{\Delta_{W}^{l(:,:)},\Delta_{X}^l} \mathit{distance}(X^{l+1}, X^{l+1}_{fp}).
\end{equation}

For layerwise quantization, there are only two scaling factors for a layer, which can be efficiently determined by simple enumerative search~\cite{EasyQuant_arxiv2020}.
For the channelwise quantization, a scaling factor affects the result of only one output channel.
So the scaling factor for each channel can be determined separately~\cite{Low-bit_quant_iccvw2019,PTQ_4bit_rapid_deployment_nips2019}.
Different from the channelwise quantization, an output channel is related to $\#H$ scaling factors in sub-layerwise quantization.
Therefore, multiple scaling factors should be jointly optimized.
However, the joint search space is exponentially increased with the increase $\#H$.
It is impractical to directly enumerate the huge space.

To solve this problem, we propose to iteratively search the scaling factors of sub-matrices as shown in Algorithm~\ref{iterative_search_scale}.
Function $\mathit{space}(\alpha,\beta,\Delta)$ generates a search space by linearly dividing interval of $[\alpha \Delta, \beta \Delta]$ into $n$ candidates. In our experiments, we set $\alpha=0.5, \beta=1.5, n=100$ and $\#\mathit{Iter}=2$.
The scaling factor is initialized to cover the maximum value in each sub-matrix, which is $\frac{\mathit{max}(\mathit{abs}({W}^{l(v,h)}))}{2^{k-1}}$.
The function $\mathit{abs}$ calculates the absolute value.
Then, we iteratively search the optimal scaling factor of each sub-matrix while fixing the other scaling factors.
Using this greedy algorithm, we significantly decrease the search difficulty.

\section{Channel Reordering}
Different from the channelwise quantization, it is possible to improve the performance of sub-layerwise quantization by reordering the weights of different channels.
In this section, we observe the potential of channel reordering then propose a method to search for the optimal order of channels.

\subsection{Potential for Channel Reordering}

\begin{figure}[tb] 
    \centering
    \includegraphics[width=0.4\textwidth]{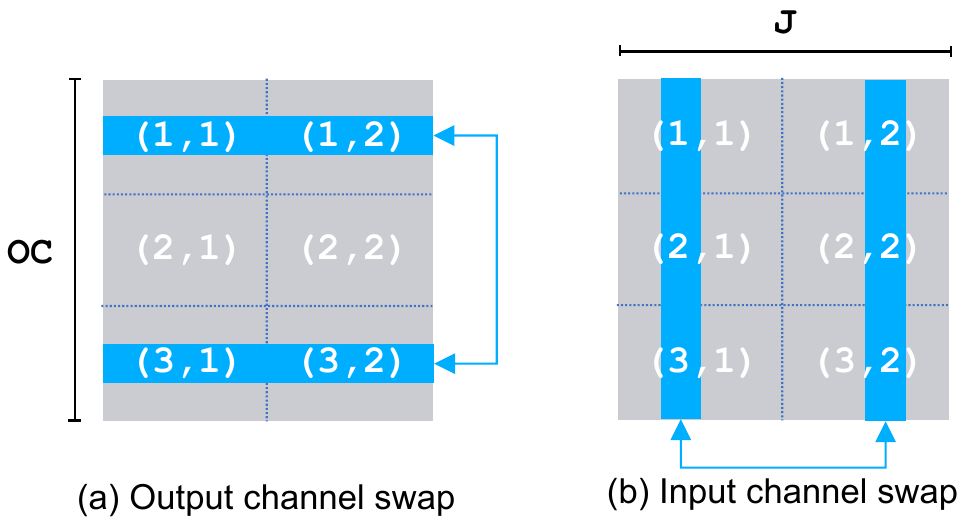} 
    \caption{Input channel swap and output channel swap. Note that an input channel may correspond to multiple columns in the weight matrix after im2col.}
    \label{im_channel_swap}
\end{figure}

Figure~\ref{im_channel_swap} demonstrates the channel reordering by swapping output channels or input channels in the weight matrix.
The swapping can change the weights located in different sub-matrices.
Therefore, we can reorder the channels to make the weights in the sub-matrix fit well with each other, which results in a lower quantization error.
It is potential to improve the performance of the sub-layerwise quantization by appropriate reordering.


\begin{figure}[tb] 
    \centering
    \includegraphics[width=0.49\textwidth]{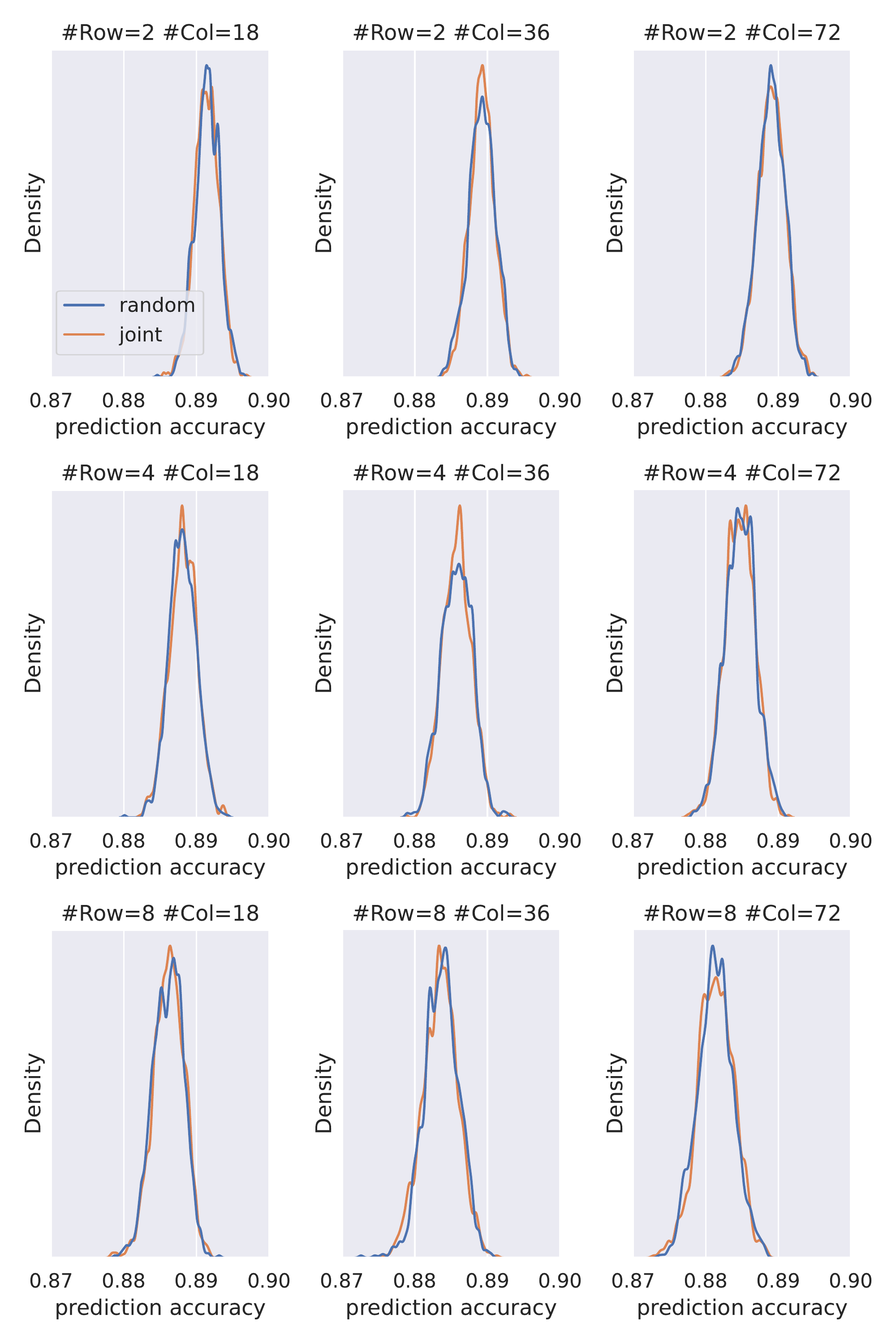} 
    \caption{The distribution of the prediction accuracy of 4-bit quantized ResNet-20 with different configurations of quantization granularity. }
    \label{resnet20_reordering}
\end{figure}

To study the channel reordering, we randomly reorder the channels of each convolutional layer in ResNet-20~\cite{He2016ResNet} and execute the sub-layerwise quantization~\footnote{Details of this experiment are described in the appendix.}.
We evaluate the prediction accuracy for different reorderings and depict the results in Figure~\ref{resnet20_reordering} (blue lines).
We observe that the prediction accuracy varies within a wide range, which indicates there exists a high potential to improve the sub-layerwise quantization through channel reordering.

\subsection{Joint Reordering in Adjacent layers}
Assume the output of the $l$-th convolutional layer is the input of the $l$+1-th convolutional layer.
If we reorder the channels of the two layers separately, it is required to rearrange the memory of feature maps $X^{l+1}$ to ensure the $l$+1-th layer gets the correct input.
However, the memory rearrangement increases the cost of inference.

To solve this problem, we propose to jointly reorder the channels in adjacent layers.
We keep the ordering of the input channels in the $l$+1-th layer the same as the ordering of the output channels in the $l$-th layer.
In this way, we can avoid memory rearrangement.
In order to understand how much this method affects the potential of reordering, we experiment on ResNet-20 using joint reordering for adjacent layers.
As shown in Figure~\ref{resnet20_reordering} (orange lines), the distributions of the prediction accuracy with joint reordering are almost the same as that with random reordering.
This indicates the joint reordering is effective.

\subsection{Search for Optimal Reordering}
Note that it is infeasible to directly evaluate the prediction accuracy in post-training quantization.
To search for the optimal reordering for each layer, we should use another metric to evaluate different reorderings.
We use the negative of the Euclidean distance between the output feature maps before and after quantization as the metric.
Because joint reordering makes the sequential layers dependent, we cannot execute the search process layer-by-layer.
However, search for the reorderings of all layers results in a huge search space, which is intractable.
Therefore, we split the network into segments and search for the optimal reorderings of the layers in each segment separately.
In our experiments, we split the models into residual blocks.
Because each block has only 2 or 3 convlutional layers, this dramatically reduces the search space.

We use the evolutionary algorithm (EA) to search for the optimal reorderings of layers in a segment.
We generate the initial population with different reorderings randomly and repeat the following evolutionary steps:
1. Evaluate the score of each reordering in the population (the negative of the distance).
2. Select the reorderings with top 50\% scores for reproduction.
3. Breed new reorderings through mutation (randomly swap some channels).
4. Replace the reorderings with the lowest scores in the population with new ones.

\section{Experiments}

In this section, we will systematically explore the sub-layerwise quantization with the proposed post-training quantization method.

\subsection{Experimental Settings}
To verify the effectiveness of sub-layerwise quantization, we select different computer vision tasks, including 1.image classification on ImageNet~\cite{imagenet_IJCV2015}, 2.object detection on COCO~\cite{mscoco_eccv2014}.
For the network, we select the widely used ResNet-18 and ResNet-50~\cite{He2016ResNet} for image classification and YOLOv3~\cite{yolov3_arxiv2018} for object detection.
As a common practice~\cite{Google_whitepaper_arxiv2018}, we fold the Batch-Normalization (BN) layer into the adjacent convolutional layer to reduce the inference overhead.
We quantize the network with $4$ bits for weights and $8$ bits for feature maps with symmetric uniform quantization.
The first layer and the last layer are not quantized.

The proposed method belongs to post-training quantization (PTQ).
We randomly sample 128 images from the training dataset for the calibration of the scaling factors.
At first, we inference the sampled images with the original floating-point network and collect the output feature maps of each layer.
Then we inference again to determine the scaling factors for weights and activations layer by layer.
For each layer, we execute the following steps.
1. We only quantize input feature maps and search for the optimal scaling factor of inputs to minimize the Euclidean distance in Eq~\ref{eq_distance}.
2. We search for the optimal scaling factor of each weight sub-matrix with Algorithm~\ref{iterative_search_scale}  fixing the scaling factor of input.
3. We search for the optimal scaling factor of inputs again fixing the scaling factors of the weights.
4. The weights and the input feature maps are quantized with the optimal scaling factors and the output is calculated as Eq~\ref{sub_layerwise_computation}.  

\subsection{Result on Image Classification}

\begin{table}[tb]
\centering
\caption{Sub-layerwise quantization for ResNet-18 with 4-bit weight and 8-bit activation.}
\label{resnet18_fixed_result}
\resizebox{\linewidth}{!}{
\setlength{\tabcolsep}{1.8mm}{
\begin{tabular}{ccccccc}
                                                           \\ \toprule
\multirow{2}{*}{$\#\mathit{Row}$} & \multicolumn{5}{c}{$\#\mathit{Col}$}   \\ \cmidrule(r){2-6}
       & 576     & 288     & 144     & 72      & 36      \\ \midrule
16     & 59.48\% & 62.21\% & 63.11\% & 64.24\% & 64.83\% \\
8      & 62.54\% & 63.80\% & 64.02\% & 65.17\% & 66.77\% \\
4      & 64.38\% & 65.42\% & 65.91\% & 66.59\% & 68.17\% \\
2      & 66.07\% & 66.55\% & 67.40\% & 68.38\% & 69.04\% \\
1      & 67.51\% & 68.10\% & 68.76\% & 69.20\% & 69.42\% \\\bottomrule
\end{tabular}}}
\end{table}

\begin{table}[tb]
\caption{Sub-layerwise quantization for ResNet-50 with 4-bit weight and 8-bit activation.}
\label{resnet50_fixed_result}
\resizebox{\linewidth}{!}{
\setlength{\tabcolsep}{1.8mm}{
\begin{tabular}{ccccccc}
                                                           \\ \toprule
\multirow{2}{*}{$\#\mathit{Row}$} & \multicolumn{5}{c}{$\#\mathit{Col}$}   \\ \cmidrule(r){2-6}
       & 576     & 288     & 144     & 72      & 36      \\ \midrule
16 & 65.93\% & 66.77\% & 68.62\% & 69.85\% & 70.30\% \\
8  & 68.56\% & 69.66\% & 70.44\% & 70.56\% & 71.07\% \\
4  & 70.58\% & 71.05\% & 71.39\% & 71.56\% & 73.30\% \\
2  & 72.40\% & 72.32\% & 72.43\% & 73.81\% & 75.27\% \\
1  & 73.39\% & 73.76\% & 74.61\% & 75.36\% & 75.70\% \\ \bottomrule
\end{tabular}}}
\end{table}

We experiment on the ResNet-18 (original precision=69.76\%) and ResNet-50 (original precision=76.13\%) on ImageNet dataset. 
To systematically study the sub-layerwise quantization, we use two methods to configure of quantization granularity.
Method 1: All layers share the same $\#\mathit{Row}$ and the same $\#\mathit{Col}$.
Method 2: The $\#\mathit{Col}$ is set by dividing the number of weights for one output channel $\#\mathit{Col}=\frac{J}{\#H}$. 

For Method 1, we select $\#\mathit{Row}\in{[1,2,4,8,16]}$ and $\#\mathit{Col}\in{[36,72,144,288,576]}$. 
We evaluate the prediction accuracy of the quantized network with different combinations of $\#\mathit{Row}$ and $\#\mathit{Col}$.
The results are shown in Table~\ref{resnet18_fixed_result} for ResNet-18 and Table~\ref{resnet50_fixed_result} for ResNet-50.
We observe that the left top point with large granularity ($\#\mathit{Row} \times \#\mathit{Col}$) has low prediciton accuracy.
While the right bottom point with small granularity has high prediction accuracy.
For example, $(16,576)$ has an accuracy of 59.48\% and $(8,288)$ is 63.80\% for ResNet-18~\footnote{For simplicity, we use $(\#\mathit{Row},\#\mathit{Col})$ to represent the configuration of quantization granularity for Method 1 and use $(\#\mathit{Row},\#H)$ for Method 2.}.
For ResNet-50, $(16,576)$ has an accuracy of 65.93\% and $(8,288)$ is 69.66\%.

\begin{table}[tb]
\caption{Sub-layerwise quantization for ResNet-18 with 4-bit weight and 8-bit activation.}
\label{resnet18_div_result}
\resizebox{\linewidth}{!}{
\setlength{\tabcolsep}{1.8mm}{
\begin{tabular}{ccccccc}
                                                           \\ \toprule
\multirow{2}{*}{$\#\mathit{Row}$} & \multicolumn{5}{c}{$\#H$}   \\ \cmidrule(r){2-6}
     & 1       & 2       & 4       & 8       & 16      \\ \midrule
16 & 56.95\% & 59.53\% & 61.70\% & 63.70\% & 65.10\% \\
8  & 59.26\% & 62.12\% & 63.57\% & 65.17\% & 65.97\% \\
4  & 62.62\% & 64.62\% & 65.49\% & 66.38\% & 67.03\% \\
2  & 64.81\% & 65.85\% & 66.71\% & 67.39\% & 68.20\% \\
1  & 66.82\% & 67.25\% & 68.10\% & 68.59\% & 68.99\% \\
\bottomrule
\end{tabular}}}
\end{table}

\begin{table}[tb]
\caption{Sub-layerwise quantization for ResNet-50 with 4-bit weight and 8-bit activation.}
\label{resnet50_div_result}
\resizebox{\linewidth}{!}{
\setlength{\tabcolsep}{1.8mm}{
\begin{tabular}{ccccccc}
                                                           \\ \toprule
\multirow{2}{*}{$\#\mathit{Row}$} & \multicolumn{5}{c}{$\#H$}   \\ \cmidrule(r){2-6}
     & 1       & 2       & 4       & 8       & 16      \\ \midrule
16 & 63.26\% & 67.36\% & 69.62\% & 71.22\% & 72.51\% \\
8  & 67.96\% & 69.58\% & 71.47\% & 72.54\% & 73.17\% \\
4  & 70.01\% & 71.79\% & 72.64\% & 73.38\% & 74.01\% \\
2  & 72.13\% & 73.17\% & 73.58\% & 74.20\% & 74.97\% \\
1  & 73.72\% & 74.17\% & 74.65\% & 75.16\% & 75.63\% \\
\bottomrule
\end{tabular}}}
\end{table}

For Method 2, we select the horizontal numbers of sub-matrices $\#H\in[1,2,4,8,16]$ and  $\#\mathit{Row}\in[36,72,144,288,576]$.
Note that the $\#\mathit{Row}\times\#\mathit{Col}$ in different layers can be different.
The results are demonstrated in Table~\ref{resnet18_div_result} for ResNet-18 and Table~\ref{resnet50_div_result} for ResNet-50.
Note that the left bottom point with configuration of $(1,1)$ is the channelwise quantization.
With the $\#H$ increase, the prediction accuracy boost.
For example, $(1,1)$ has an accuracy of 66.82\% and $(1,2)$ is 67.25\% for ResNet-18.
For ResNet-50, $(1,1)$ has an accuracy of 73.72\% and $(1,2)$ is 74.17\%.

\subsection{Result on Object Detection}
\begin{table}[tb]
\centering
\caption{Sub-layerwise quantization for YOLOv3-320 with 4-bit weight and 8-bit activation. The metric is mAP.}
\label{yolov3_320_result}
\resizebox{\linewidth}{!}{
\setlength{\tabcolsep}{3mm}{
\begin{tabular}{cccccc}
\toprule
\multirow{2}{*}{$\#\mathit{Row}$} & \multicolumn{4}{c}{$\#\mathit{Col}$}   \\ \cmidrule(r){2-6}
                         & 576   & 288   & 144   & 72     & 36    \\
\midrule
16                       & 0.220 & 0.222 & 0.232 & 0.240  & 0.249 \\
8                        & 0.234 & 0.235 & 0.240 & 0.245  & 0.258 \\
4                        & 0.244 & 0.244 & 0.253 & 0.259  & 0.266 \\
2                        & 0.255 & 0.256 & 0.262 & 0.269  & 0.271 \\
1                        & 0.266 & 0.268 & 0.273 & 0.274  & 0.276 \\
\bottomrule
\end{tabular}}}
\end{table}

Then we evaluate the sub-layerwise quantization on the object detection task.
Table~\ref{yolov3_320_result} demonstrates the result of YOLOv3-320 (original mAP=0.282) on COCO dataset.
The results are similar to ResNet-18 and ResNet-50.

\subsection{Correlation between Accuracy and Granularity}
From the above experiments, we conclude that the prediction accuracy of the quantized network has a strong correlation with the quantization granularity.
Intuitively, a small quantization granularity results in a smaller number of weights in each sub-matrix and a higher possibility they fit well with each other.
For example, if we quantize the matrix $\left[\begin{smallarray}{cc}0.1 & -0.8\\0.5&-1.5\end{smallarray}\right]$ to 4-bit with a single scaling factor, any scaling factor results in quantization error. 
If we use sub-layerwise quantization $\#\mathit{Row}=1$ and $\#\mathit{Col}=2$, it is easy to achieve zero quantization error.

We observe that the configuration with lower $\#\mathit{Row}$ has better prediction accuracy when two configurations have the same $\#\mathit{Row} \times \#\mathit{Col}$.
For example, the point $(1,1)$ has an accuracy of 66.82\% while $(2,2)$ is 65.85\% for ResNet-18 in Table~\ref{resnet18_div_result}.
For ResNet-50, the point $(1,1)$ has an accuracy of 73.72\% while $(2,2)$ is 73.17\% in Table~\ref{resnet50_div_result}.
This indicates that the distributions of the weights in different output channels are different.
The weights with diverse distribution cannot fit well with each other, increasing the quantization error if they are quantized with the same scaling factor.
Because we fold the BN into the convolutional layer, it increases the difference of the distributions of the weights in different channels.
To verify this, we experiment on the ResNet-18 and ResNet-50 without folding the BN into the convolutional layer.
As shown in Table~\ref{no_fold}, we set the same $\#\mathit{Row} \times \#\mathit{Col}$ with different configurations.
The difference of different configurations is much smaller than that with BN folded.
Intuitively, we can further reduce the distribution difference by reordering to make the weights with similar distribution in a group.

\begin{table}[tb]
\caption{Sub-layerwise quantization without folding the BN into the convolutional layer. }
\label{no_fold}
\resizebox{\linewidth}{!}{
\setlength{\tabcolsep}{1.05mm}{
\begin{tabular}{ccccccc}
                                                          \\ \toprule
\multirow{2}{*}{$\mathit{Net}$} & \multicolumn{5}{c}{($\#\mathit{Row}$,$\#H$)}   \\ \cmidrule(r){2-6}
          & (16, 16) & (8,8)   & (4,4)   & (2,2)   & (1,1)   \\ \midrule
ResNet-18 & 66.60\%  & 66.15\% & 66.53\% & 67.13\% & 67.54\% \\
ResNet-50 & 72.99\%  & 73.38\% & 73.54\% & 73.58\% & 73.74\% \\ \bottomrule
\end{tabular}}}
\end{table}

\subsection{Result of Channel Reordering}
In order to further improve the sub-layerwise quantization, we jointly search for the optimal reorderings in each residual block using an evolutionary algorithm.
After each reordering of channels, we calibrate the scaling factors of the weights and activations in the block.
And we use the negative of the Euclidean distance between the outputs of the block before and after as the score for the evolutionary algorithm.
The number of individuals in the population is set to 40 and the number of evolution iterations is set to 5.

Table~\ref{resnet18_reordering} demonstrates the result on ResNet-18~\footnote{More results are presented in the appendix.}.
The improvement of the prediction accuracy is from 0.10\% to 2.55\%.
We observe that the left top point has a large improvement.
This is because that the larger granularity, the higher probability that weights with very different distributions will be placed in a group.
Reordering can place them in different groups, which dramatically reduces the quantization error.
We also find that there is randomness in performance improvement, such as the significant improvement of (2,72).
The randomness comes from the heuristic search method, which not fully explore the search space.
There is potential to further increase the performance.

From the results, we find that a lot of points outperform the channelwise quantization (ResNet-18 is 66.82\%, ResNet-50 is 73.72\% and YOLOv3-320 is 0.266) in both experiments with reordering and experiments without reordering.
This indicates that the channelwise quantization is not the optimal option considering only the prediction accuracy.

\begin{table}[tb]
\centering
\caption{Result of channel reordering for ResNet-18. The percentages in parentheses are the increment comparing with results in Table~\ref{resnet18_fixed_result}.}
\label{resnet18_reordering}
\resizebox{0.96\linewidth}{!}{
\setlength{\tabcolsep}{2mm}{
\begin{tabular}{ccccc}
\toprule
\multirow{2}{*}{$\mathit{\#Row}$} & \multicolumn{4}{c}{$\mathit{\#Col}$}          \\ \cmidrule(r){2-5} 
                                  & 576       & 288       & 144       & 72        \\
\midrule
          \multirow{2}{*}{16}     & 62.03\%   & 63.24\%   & 64.44\%   & 64.84\%   \\
                                  & (2.55\%)  & (1.03\%)  & (1.33\%)  & (0.59\%)  \\
          \multirow{2}{*}{8}      & 63.60\%   & 64.43\%   & 65.13\%   & 65.78\%   \\
                                  & (1.07\%)  & (0.63\%)  & (1.12\%)  & (0.60\%)  \\
          \multirow{2}{*}{4}      & 64.90\%   & 65.79\%   & 66.29\%   & 68.35\%   \\
                                  & (0.52\%)  & (0.37\%)  & (0.38\%)  & (1.76\%)  \\
          \multirow{2}{*}{2}      & 66.27\%   & 67.09\%   & 67.87\%   & 68.78\%   \\
                                  & (0.20\%)  & (0.54\%)  & (0.48\%)  & (0.39\%)  \\
          \multirow{2}{*}{1}      & 67.71\%   & 68.27\%   & 68.98\%   & 69.31\%   \\
                                  & (0.20\%)  & (0.18\%)  & (0.22\%)  & (0.10\%)  \\
\bottomrule
\end{tabular}}}
\end{table}

\section{Discussion}
\label{discussion}
As sub-layerwise quantization is not the free lunch, in this section, we will discuss the computation overhead and the memory overhead.
\subsection{Computation Overhead}

Looking back at Eq~\ref{sub_layerwise_computation}, we find that there are extra computation that the scaling factors $\Delta_{W}^{l(v,h)} \Delta_{X}^l$ multiply the result of sub-matrix multiplication $\sum_{j} Q(W^{l(v,h)}_{cj})Q(X_{jp}^{l(h)})$.
The number of Multiply-Accumulations (MACs) of the extra computation is $\#H\times\mathit{OC}\times\mathit{P}$, where the $\mathit{OC}$ is the number of output channels and the $\mathit{P}$ is the number of pixels in the output feature map.
The number of MACs of a normal convolutional layer is $\mathit{OC}\times\mathit{P}\times{K}^2\times\mathit{IC}$, where the $K^2$ is the kernel size and the $\mathit{IC}$ is the number of input channels.
The computation overhead can be calculated by ${\frac{\#H\times\mathit{OC}\times\mathit{P}}{\mathit{OC}\times\mathit{P}\times{K}^2\times\mathit{IC}}=\frac{\#H}{K^2\times \mathit{IC}}}$

\begin{figure}[tb] 
    \centering
    \includegraphics[width=0.45\textwidth]{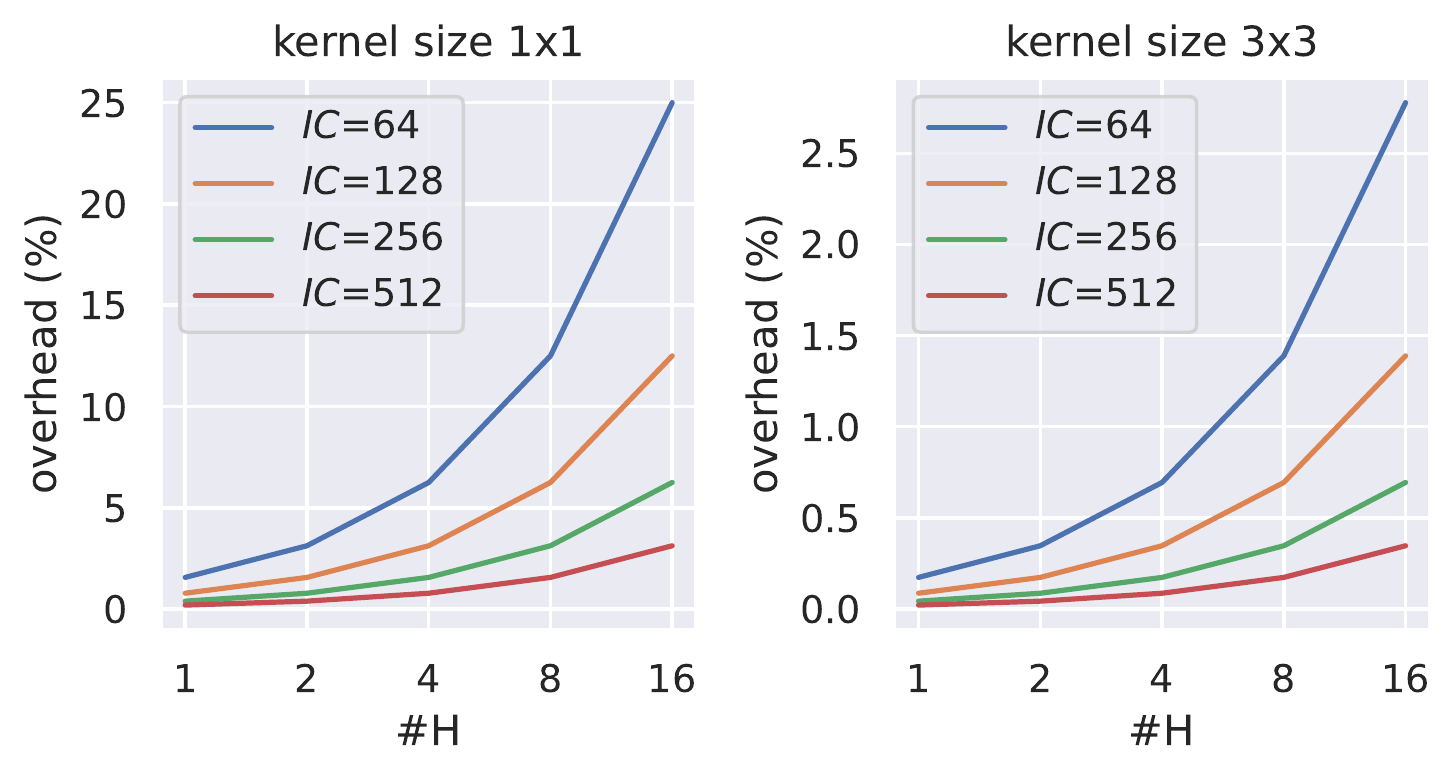} 
    \caption{The computation overhead for the sub-layerwise quantization with different configurations for the convolutional layer. $\mathit{IC}$ is the number of input channels.}
    \label{im_overhead}
\end{figure}

To analyze the extra computation, we plot the computation overhead in Figure~\ref{im_overhead} with different configurations.
We observe that the overhead is greatly affected by the kernel size. 
The overhead can be low for the 3x3 kernel size and larger kernel size.
Although 1x1 kernel size has up to 25\% computation overhead, large overhead only occurs in a small number of layers.



\begin{table}[tb]
\caption{The computation overhead of Sub-layerwise quantization for different networks.}
\label{resnet_overhead}
\resizebox{\linewidth}{!}{
\begin{tabular}{@{}cccccc@{}}
\toprule
$\#\mathit{Col}$       & 576    & 288    & 144    & 72     & 36     \\ \midrule
ResNet-18 & 0.20\% & 0.37\% & 0.73\% & 1.42\% & 2.79\%    \\
ResNet-50 & 0.29\% & 0.41\% & 0.72\% & 1.35\% & 2.65\%    \\
YOLOv3    & 0.21\% & 0.37\% & 0.72\% & 1.41\% & 2.80\%    \\
\midrule
$\#\mathit{V}$       & 1    & 2    & 4    & 8     & 16    \\ \midrule
ResNet-18 & 0.13\% & 0.25\% & 0.51\% & 1.02\% & 2.04\% \\
ResNet-50 & 0.24\% & 0.47\% & 0.94\% & 1.88\% & 3.76\% \\ 
YOLOv3    & 0.12\% & 0.24\% & 0.48\% & 0.95\% & 1.90\% \\            
\bottomrule
\end{tabular}}
\end{table}

We calculated the computation overhead of the ResNet and YOLOv3 for different configurations of quantization granularity.
As shown in Table~\ref{resnet_overhead}, the computation overhead increases as $\#\mathit{Col}$ decreases and $\#V$ increases.
We observe that the maximum computation overhead is 3.76\%. 

\subsection{Memory Overhead}
For the memory footprint, the overhead is to save the scaling factors.
The number of scaling factors is $\#\mathit{V}\times\#\mathit{H}$ and the number of weights is $\mathit{OC}\times J$.
Because $\#\mathit{V}=\lceil \frac{OC}{\#\mathit{Col}}\rceil$ and $\#\mathit{H}=\lceil \frac{J}{\#\mathit{Row}}\rceil$, the memory overhead can be approximated by $\frac{\#\mathit{V}\times\#\mathit{H}}{\mathit{OC}\times J} \approx \frac{1}{\#\mathit{Col}\times \#\mathit{Row}}$.
The memory overhead is inversely proportional to $\#\mathit{Col}\times \#\mathit{Row}$.
We also calculated the memory overhead of the ResNet and YOLOv3 with different configurations of quantization granularity.
We observe that most of the overhead is less than 1\% and the maximum one is 2.8\%.
Both the computation overhead and memory overhead is not significant comparing with the channelwise quantization.
Considering the performance and insignificant overhead, sub-layerwise quantization can be a better choice than channelwise quantization.





\section{Related Work}

Various quantization methods have been proposed to compress the convolutional neural networks.
The training-aware quantization methods~\cite{PACT_arxiv2018,LQNets_eccv2018,Quantization_training_integer_only_cvpr2018,Google_whitepaper_arxiv2018} use the training dataset to fine-tune the quantized network through back-propagation.
While the post-training quantization methods~\cite{PTQ_4bit_rapid_deployment_nips2019,Low-bit_quant_iccvw2019,Adaround_icml2020,Adaquant_arxiv2020,BRECQ_ICLR2021} quantize the network with some unlabeled calibration images.

There are various quantization granularities.
Layerwise quantization quantizes the weight in a layer with a single scaling factor.
While sub-layerwise quantization quantizes different groups of weights in a layer with different scaling factors.
The channelwise quantization is a case of sub-layerwise quantization and it is widely used in different quantization methods ~\cite{Survey_quantization_arxiv2021}. 
Other sub-layerwise granularities are not well explored.
\cite{Bit_goes_down_reivisiting_quantization_iclr2020} proposed the sub-channelwise quantization. 
It divides each channel into several groups.
Q-BERT~\cite{Q-BERT_aaai2020} proposed to quantize the weights sub-layerwisely in Transformer.
Sub-layerwise quantization-aware training is used in \cite{Low_bitwidth_CNN_RRAM_TCAD2020}, which splits a single convolutional layer into multiple convolutional layers according to the granularity.

\section{Conclusion}


In this paper, we proposed an efficient sub-layerwise post-training quantization method (PTQ-SL) and systematically explored the quantization granularity.
From the experiment, we concluded that the prediction accuracy of the quantized neural network has a strong correlation with granularity.
We also observed that there is a great potential to improve the performance of quantization by channel reordering.
The experiments demonstrated that the sub-layerwise quantization with appropriate channel reordering can outperform the channelwise quantization.
Then, we discussed the computation overhead and memory overhead of different quantization granularity.
We concluded that sub-layerwise quantization can be a better choice than channelwise quantization.
We look forward to further exploring sub-layerwise quantization to improve the quantization performance in the future.

\section{Acknowledgments}
This work is supported by National Key R\&D Program of China (2020AAA0105200) and National Natural Science Foundation of China (61832020, 62032001, 92064006).

\bibliography{aaai22}


\section{Appendix}

\subsection{Environment of Experiments}
We execute the experiments on a server with 6 Nvidia Tesla V100 cards.
The operating system is Ubuntu 20.04.3.
The deep learning framework is Pytorch 1.9.0~\cite{NEURIPS2019_9015}.
We use ResNet-18 and ResNet-50 provided by torchvision 0.10.0.
We use YOLOv3-320 provided by MMDetection 2.14.0~\cite{mmdetection}.

\subsection{The Settings of Experiments on ResNet-20}


To demonstrate the difference with and without joint reordering, we experiment on a small network.
ResNet-20 is a network for the CIFAR-10 image classification task.
We use the ResNet-20 provided by~\cite{Idelbayev18a} and the original prediction accuracy of the network is 91.73\%.
We set the number of calibration samples to 64 and select $\#\mathit{Row}\in{[2,4,8]}$ and $\#\mathit{Col}\in{[18,36,72]}$.
Other calibration parameter settings are the same as the experiment section.

For each configuration of quantization granularity, We randomly sample different reorderings for 1,000 times.
In each residual block, we reorder the output channels of the first convolutional layer and the input channels of the second convolutional layer.
For each reordered network, we perform sub-layerwise post-training quantization and test the predicition accuracy of the quantized networks.
Both weights and activations are quantized to 4-bit with sub-layerwise quantization.
Then we plot the distribution of the prediction accuracy as shown in Figure~\ref{resnet20_reordering}.

\subsection{Results of Object Detection}

\begin{table}[tb]
\centering
\caption{Results of YOLOv3-320 on COCO for different configurations of sub-layerwise quantization.}
\label{detection_fixed}
\resizebox{\linewidth}{!}{
\setlength{\tabcolsep}{2mm}{
\begin{tabular}{ccccccccc}
\toprule
\multirow{2}{*}{metric}                   & \multirow{2}{*}{$\mathit{\#Rows}$} & \multicolumn{5}{c}{{$\mathit{\#Cols}$}}            \\
                                          &                         & 576   & 288   & 144   & 72    & 36    \\
\midrule
\multirow{5}{*}{$\textup{AP}$}            & 16                      & 0.221 & 0.222 & 0.232 & 0.240 & 0.249 \\
                                          & 8                       & 0.233 & 0.234 & 0.240 & 0.245 & 0.258 \\
                                          & 4                       & 0.245 & 0.246 & 0.253 & 0.259 & 0.266 \\
                                          & 2                       & 0.254 & 0.258 & 0.262 & 0.269 & 0.271 \\
                                          & 1                       & 0.265 & 0.268 & 0.271 & 0.274 & 0.276 \\ \midrule
\multirow{5}{*}{$\textup{AP}^{0.5}$}      & 16                      & 0.400 & 0.408 & 0.423 & 0.432 & 0.449 \\
                                          & 8                       & 0.418 & 0.423 & 0.436 & 0.447 & 0.462 \\
                                          & 4                       & 0.440 & 0.443 & 0.455 & 0.454 & 0.473 \\
                                          & 2                       & 0.456 & 0.461 & 0.468 & 0.476 & 0.479 \\
                                          & 1                       & 0.469 & 0.472 & 0.477 & 0.483 & 0.485 \\\midrule
\multirow{5}{*}{$\textup{AP}^{0.75}$}     & 16                      & 0.217 & 0.217 & 0.225 & 0.238 & 0.246 \\
                                          & 8                       & 0.229 & 0.232 & 0.239 & 0.240 & 0.254 \\
                                          & 4                       & 0.244 & 0.244 & 0.251 & 0.258 & 0.269 \\
                                          & 2                       & 0.252 & 0.256 & 0.261 & 0.271 & 0.274 \\
                                          & 1                       & 0.266 & 0.270 & 0.275 & 0.275 & 0.281 \\\midrule
\multirow{5}{*}{$\textup{AP}_{s}$}        & 16                      & 0.071 & 0.062 & 0.068 & 0.072 & 0.078 \\
                                          & 8                       & 0.072 & 0.064 & 0.070 & 0.076 & 0.087 \\
                                          & 4                       & 0.077 & 0.074 & 0.083 & 0.093 & 0.088 \\
                                          & 2                       & 0.082 & 0.087 & 0.090 & 0.101 & 0.097 \\
                                          & 1                       & 0.091 & 0.100 & 0.098 & 0.103 & 0.103 \\\midrule
\multirow{5}{*}{$\textup{AP}_{m}$}        & 16                      & 0.217 & 0.220 & 0.233 & 0.246 & 0.262 \\
                                          & 8                       & 0.240 & 0.241 & 0.247 & 0.252 & 0.272 \\
                                          & 4                       & 0.257 & 0.259 & 0.266 & 0.270 & 0.287 \\
                                          & 2                       & 0.263 & 0.271 & 0.280 & 0.288 & 0.290 \\
                                          & 1                       & 0.282 & 0.282 & 0.290 & 0.290 & 0.297 \\\midrule
\multirow{5}{*}{$\textup{AP}_{l}$}        & 16                      & 0.378 & 0.385 & 0.395 & 0.403 & 0.408 \\
                                          & 8                       & 0.386 & 0.395 & 0.404 & 0.411 & 0.423 \\
                                          & 4                       & 0.403 & 0.408 & 0.415 & 0.421 & 0.427 \\
                                          & 2                       & 0.413 & 0.417 & 0.423 & 0.429 & 0.428 \\
                                          & 1                       & 0.422 & 0.424 & 0.427 & 0.432 & 0.432 \\ \bottomrule

\end{tabular}}}
\end{table}

\begin{table}[tb]
\centering
\caption{Results of YOLOv3-320 on COCO for different configurations of sub-layerwise quantization.}
\label{detection_divide}
\resizebox{\linewidth}{!}{
\setlength{\tabcolsep}{2mm}{
\begin{tabular}{ccccccccc}
\toprule
\multirow{2}{*}{metric}                   & \multirow{2}{*}{$\mathit{\#Rows}$} & \multicolumn{5}{c}{{$\mathit{\#H}$}}            \\
                                          &                         & 1     & 2     & 4     & 8     & 16    \\ \midrule
\multirow{5}{*}{$\textup{AP}$}            & 16 & 0.219 & 0.220 & 0.222 & 0.231  & 0.241 \\
                                          & 8  & 0.227 & 0.233 & 0.231 & 0.243  & 0.248 \\
                                          & 4  & 0.237 & 0.241 & 0.248 & 0.254  & 0.259 \\
                                          & 2  & 0.249 & 0.254 & 0.258 & 0.262  & 0.268 \\
                                          & 1  & 0.261 & 0.264 & 0.268 & 0.273  & 0.274 \\
\midrule
\multirow{5}{*}{$\textup{AP}^{0.5}$}      & 16 & 0.398 & 0.402 & 0.405 & 0..416 & 0.434 \\
                                          & 8  & 0.410 & 0.418 & 0.421 & 0.440  & 0.453 \\
                                          & 4  & 0.426 & 0.436 & 0.444 & 0.456  & 0.467 \\
                                          & 2  & 0.446 & 0.457 & 0.463 & 0.470  & 0.477 \\
                                          & 1  & 0.464 & 0.471 & 0.475 & 0.482  & 0.483 \\ \midrule
\multirow{5}{*}{$\textup{AP}^{0.75}$}     & 16 & 0.215 & 0.215 & 0.218 & .0.230 & 0.240 \\
                                          & 8  & 0.224 & 0.231 & 0.229 & 0.240  & 0.243 \\
                                          & 4  & 0.236 & 0.238 & 0.247 & 0.252  & 0.256 \\
                                          & 2  & 0.249 & 0.254 & 0.259 & 0.262  & 0.271 \\
                                          & 1  & 0.261 & 0.266 & 0.268 & 0.276  & 0.271 \\ \midrule
\multirow{5}{*}{$\textup{AP}_{s}$}        & 16 & 0.069 & 0.071 & 0.063 & 0.057  & 0.071 \\
                                          & 8  & 0.069 & 0.077 & 0.063 & 0.072  & 0.080 \\
                                          & 4  & 0.077 & 0.079 & 0.082 & 0.083  & 0.090 \\
                                          & 2  & 0.081 & 0.090 & 0.091 & 0.094  & 0.098 \\
                                          & 1  & 0.089 & 0.091 & 0.096 & 0.103  & 0.103 \\ \midrule 
\multirow{5}{*}{$\textup{AP}_{m}$}        & 16 & 0.219 & 0.216 & 0.221 & 0.240  & 0.247 \\
                                          & 8  & 0.231 & 0.238 & 0.236 & 0.253  & 0.258 \\
                                          & 4  & 0.245 & 0.245 & 0.259 & 0.269  & 0.272 \\
                                          & 2  & 0.257 & 0.263 & 0.271 & 0.277  & 0.288 \\
                                          & 1  & 0.273 & 0.279 & 0.286 & 0.292  & 0.295 \\ \midrule
\multirow{5}{*}{$\textup{AP}_{l}$}        & 16 & 0.366 & 0.379 & 0.383 & 0.395  & 0.400 \\
                                          & 8  & 0.379 & 0.390 & 0.395 & 0.402  & 0.405 \\
                                          & 4  & 0.393 & 0.402 & 0.408 & 0.412  & 0.415 \\
                                          & 2  & 0.404 & 0.412 & 0.418 & 0.417  & 0.424 \\
                                          & 1  & 0.415 & 0.420 & 0.422 & 0.428  & 0.433 \\ \bottomrule
\end{tabular}}}
\end{table}

We experiment on the YOLOv3-320 on COCO 2017 objection detection task. 
The original $\textup{AP}$ is 0.280, $\textup{AP}^{0.5}$ is 0.492, $\textup{AP}^{0.75}$ is 0.283, $\textup{AP}_{s}$ is 0.105, $\textup{AP}_{m}$ is 0.301, and $\textup{AP}_{l}$ is 0.439.     
To systematically study the sub-layerwise quantization, we use two methods to configure quantization granularity.
Method 1: All layers share the same $\#\mathit{Row}$ and the same $\#\mathit{Col}$.
Method 2: The $\#\mathit{Col}$ is set by dividing the number of weights for one output channel $\#\mathit{Col}=\frac{J}{\#H}$. 

For Method 1, we evaluate the prediction accuracy of the quantized network with different combinations of $\#\mathit{Row}$ and $\#\mathit{Col}$.
The results of $\textup{AP}$, $\textup{AP}^{0.5}$, $\textup{AP}^{0.75}$, $\textup{AP}_{s}$, $\textup{AP}_{m}$, and $\textup{AP}_{l}$ are reported in Table~\ref{detection_fixed}.
We observe that the left top point with large granularity ($\#\mathit{Row} \times \#\mathit{Col}$) has low prediction accuracy.
While the right bottom point with small granularity has high prediction accuracy.
There are several points not follow this trend, such as $(2,36)$ and $(2,72)$ for metric $\textup{AP}_{l}$.

For Method 2, we select the horizontal numbers of sub-matrices $\#H\in[1,2,4,8,16]$ and  $\#\mathit{Row}\in[36,72,144,288,576]$.
The results are demonstrated in Table~\ref{detection_divide}.
Note that the left bottom point with the configuration of $(1,1)$ is the channelwise quantization.
We observe that there are also a lot of points that outperform it.

\subsection{More Results for Channel Reordering}

\begin{table}[tb]
\centering
\caption{Results of channel reordering for ResNet-50. The percentages in parentheses are the increment comparing with results without reordering.}
\label{resnet50_reordering}
\resizebox{0.9\linewidth}{!}{
\setlength{\tabcolsep}{2mm}{
\begin{tabular}{cccccc}
\toprule
\multirow{2}{*}{$\mathit{\#Rows}$} & \multicolumn{4}{c}{$\mathit{\#Cols}$}            \\
                                                     & 576     & 288     & 144     & 72      \\
\midrule
                            \multirow{2}{*}{16}     & 66.33\% & 67.28\% & 69.43\% & 70.10\% \\
                                                 & (0.40\%)& (0.51\%)& (0.81\%)& (0.24\%)\\
                            \multirow{2}{*}{8}      & 69.24\% & 69.92\% & 70.89\% & 71.06\% \\
                                                 & (0.68\%)& (0.25\%)& (0.44\%)& (0.50\%)\\
                            \multirow{2}{*}{4}      & 71.10\% & 71.23\% & 71.69\% & 72.09\% \\
                                                 & (0.52\%)& (0.18\%)& (0.30\%)& (0.53\%)\\
                            \multirow{2}{*}{2}      & 72.53\% & 72.66\% & 72.81\% & 74.41\% \\
                                                 & (0.13\%)& (0.33\%)& (0.38\%)& (0.60\%)\\
                            \multirow{2}{*}{1}      & 73.67\% & 73.99\% & 74.91\% & 75.48\% \\
                                                 & (0.28\%)& (0.23\%)& (0.29\%)& (0.12\%)\\
\bottomrule
\end{tabular}}}
\end{table}

We jointly search for the optimal reordering in each residual block using the evolutionary algorithm.
The number of individuals in the population is set to 40 and the number of evolution iterations is set to 5.
When we initialize the population, we add the individual without reordering into the population, which ensures that the result is not worse than the original.
The mutation is implemented by randomly selecting up to 30 pairs of channels and swapping them.

Table~\ref{resnet50_reordering} demonstrates the results on ResNet-50.
The improvement of the prediction accuracy is from 0.12\% to 0.81\%.
We observe more randomness in performance improvement than the results on ResNet-18.
This is because there are more layers in ResNet-50 than ResNet-18 and it has a larger search space.
Searching in a larger space is more difficult.

\end{document}